\pdfoutput=1
\PassOptionsToPackage{prologue,dvipsnames}{xcolor}

\documentclass[11pt]{article}

\usepackage[preprint]{acl}

\usepackage{times}
\usepackage{latexsym}

\usepackage{kotex}
\usepackage{multirow}
\usepackage{makecell}

\usepackage[dvipsnames]{xcolor}
\PassOptionsToPackage{dvipsnames}{xcolor}
\usepackage{booktabs}
\usepackage{amsmath}
\usepackage{adjustbox}

\usepackage[T1]{fontenc}
\usepackage[utf8]{inputenc}

\usepackage{microtype}
\usepackage{graphicx}

\usepackage{inconsolata}

\title{KoCoNovel: Annotated Dataset of Character\\ Coreference in Korean Novels}

\author{Kyuhee Kim, Surin Lee \and Sangah Lee \\
  Seoul National University \\
  \texttt{\{salgu, lsrwhite, sanalee\}@snu.ac.kr}}

\begin{document}
\maketitle
\begin{abstract}
In this paper, we present KoCoNovel, a novel character coreference dataset derived from Korean literary texts, complete with detailed annotation guidelines. Comprising 178K tokens from 50 modern and contemporary novels, KoCoNovel stands as one of the largest public coreference resolution corpora in Korean, and the first to be based on literary texts. KoCoNovel offers four distinct versions to accommodate a wide range of literary coreference analysis needs. These versions are designed to support perspectives of the omniscient author or readers, and to manage multiple entities as either separate or overlapping, thereby broadening its applicability.

One of KoCoNovel's distinctive features is that 24\% of all character mentions are single common nouns, lacking possessive markers or articles. This feature is particularly influenced by the nuances of Korean address term culture, which favors the use of terms denoting social relationships and kinship over personal names. In experiments with a BERT-based coreference model, we observe notable performance enhancements with KoCoNovel in character coreference tasks within literary texts, compared to a larger non-literary coreference dataset. Such findings underscore KoCoNovel's potential to significantly enhance coreference resolution models through the integration of Korean cultural and linguistic dynamics.\footnote{We release our data and models to the community at \url{https://github.com/storidient/KoCoNovel.git}}
\end{abstract}

\section{Introduction}
%intro-first paragraph
In narrative fiction, effective storytelling relies heavily on character actions and dialogues, making character coreference resolution critical for analyzing novels. This essential process, which maps textual references to characters throughout a narrative, faces challenges due to characters' evolving identities and relationships \citep{rosiger}. The complexity increases in stories where characters start as separate entities but are later shown to be connected, such as the revelation of a criminal's identity \citep{fantasy}. Consequently, significant research has been directed towards creating specialized datasets and guidelines for co-reference resolution in English literature (\citealp{fantasy}; \citealp{litbank}).

\begin{table}[t!]
\centering
\resizebox{7.8cm}{!}{
\begin{tabular}{ll}
\toprule
\textit{KO} & "달걀이 어디서 났니?" 하고 \{조부님\}$_a$은 물으셨다.\\
\textit{EN} & "Where did the eggs come from?" asked \textbf{\{the}\\
& \textbf{grandfather\}$_a$}.\\

\midrule
\textit{KO} & "{\{\{삼순이\}$_b$ 어머니\}}$_c$더러 {\{오빠\}}$_d$ 온단 말을 했더니,\\
& 달걀 두 개를 주어. 갖다가 쪄 주라구." \\
\textit{EN} & "I($\emptyset$) told \textbf{{\{\{Sam-soon\}$_b$'s mother\}}$_c$} that \textbf{{\{brother\}}$_d$}\\
& was coming, and she($\emptyset$) gave me($\emptyset$) two eggs to steam \\
& for him($\emptyset$)."\\

\midrule
\textit{KO} & 하고 \{누이\}$_e$가 만족한 듯이 대답한다.\\
\textit{EN} & replied \textbf{{\{the sister\}}$_e$}, seemingly pleased.\\
\midrule
\textit{KO} & "\{\{보부\}$_f$ 어미\}$_g$ 오늘도 안 왔던?"\\
\textit{EN} & "Did \textbf{{\{\{Bobu\}$_f$'s mother\}}$_g$} not come today either?" \\

\midrule
\textit{KO} & "{\{\{보부\}$_f$ 엄마\}}$_g$가 누구냐?" \\
\textit{EN} & "Who is \textbf{{\{\{Bobu\}$_f$'s mother\}}$_g$}"\\

\midrule
\textit{KO} & 하고 \{나\}$_d$는 못 듣던 \{여인\}$_g$의 이름이 수상해서\\
& \{경애\}$_e$에게 물었다.\\
\textit{EN} & \textbf{\{I\}$_d$} asked \textbf{\{Kyung-ae\}$_e$}, curious about the name of \\
& \textbf{\{the woman\}}$_g$ I($\emptyset$) had not heard before.\\
\bottomrule
\end{tabular}}
\caption{An example of coreferences in KoCoNovel, showcasing the culture of address terms in Korea. The empty set symbol($\emptyset$) is due to Korean's unique trait of omitting sentence arguments. The sentences are from \textit{The Unknown Woman} by Lee.}
\label{tab:sample_rep}
\end{table}

\begin{table*}[t!]
\centering
\resizebox{\textwidth}{!}{
\begin{tabular}{lccccccc}
\toprule
& \textbf{DROC}  
& \textbf{Litbank} 
& \textbf{FantasyCoref} 
& \textbf{OntoNotes}
& \textbf{NIKL} 
& \textbf{ETRI}
& \textbf{KoCoNovel}
\\
\midrule
Language & German & English & English & English
& Korean & Korean & Korean\\
Annotation Coverage & Character & 6 Entity Types
& All & All & All & All & Character\\
\midrule
Works (Docs) & 90 & 100 & 214 & 3493 & 7687 & 645 (4035*) & 50\\
Tokens & 393K & 210K & 367K & 1600K & 3006K & 12K & 178K\\
Tokens per Doc & 4368 & 2105 & 1719 & 466 & 391 & 19 & 3578\\
\bottomrule
\end{tabular}}
\caption{Comparison between KoCoNovel and other coreference resolution datasets: The ETRI dataset comprises 4035 documents, but only 645 documents are publicly available.}
\label{tab:stats}
\end{table*}

%intro-second paragraph
However, there is a lack of coreference datasets based on literary texts in Korean. Considering that literature encodes specific cultural aspects and reflects the unique characteristics of a language culture \citep{culture}, this gap cannot simply be bridged by translating datasets from English. In Korea, there exists a developed culture of address terms, whereby individuals are often referred to by general nouns that reveal their relationship, such as their title or kinship terms, rather than by their names \citep{kang}. This practice is far more extensive than in English, where titles primarily indicate the referent’s gender or marital status (see Table~\ref{tab:sample_rep}). Moreover, the Korean language lacks indefinite and definite articles and does not use markers to distinguish proper nouns from common nouns, such as capitalization. These cultural and linguistic characteristics underscore the necessity for a coreference dataset, based on new guidelines tailored to Korean.

%intro-third paragraph
In this study, we develop coreference guidelines tailored for Korean literary texts and introduce the KoCoNovel dataset, which consists of character coreference annotations across fifty works of modern and contemporary Korean literature. KoCoNovel stands as the first Korean coreference dataset sourced from literary texts, and it comprises 178K words, 19K character mentions, and 1.4K entities. Our contributions are as follows: 

\begin{itemize}

\item Considering the culture of address terms and the linguistic features of Korean, we revise guidelines including the criteria for span (the range of mention) (Section~\ref{sec:maxspan}) and the guidelines for distinguishing target mentions (Section~\ref{sec:versus}).

\item In addressing the diversity of styles in modern and contemporary novels influenced by Modernism, we refine the character identification criteria outlined in DROC \citep{droc} to accommodate novels of various styles (Section~\ref{sec:defchar}). 

\item Building on the current discussion of coreference resolution in literary texts, we offer four unique versions of KoCoNovel to provide a comprehensive dataset for exploration:

\textbf{[Reader/Omniscient]} From the perspective of the omniscient author or the readers (Section~\ref{sec:asym}).

\textbf{[Separate/Overlapped]} How multiple entities are treated either as separate entities (e.g., [\textquoteleft We\textquoteright], [\textquoteleft I\textquoteright], [\textquoteleft You\textquoteright]) or as overlapped entities (e.g., [\textquoteleft We\textquoteright, 
\textquoteleft I\textquoteright], [\textquoteleft We\textquoteright, 
\textquoteleft You\textquoteright]) (Section~\ref{sec:multi}).

\item By training BERT-based coreference models with the KoCoNovel dataset, we achieved significant enhancements in performance for character coreference tasks within literary texts, surpassing models trained on a larger non-literary corpus.
\end{itemize}

\section{Background}
%back-first paragraph
Several studies have produced extensive datasets for coreference resolution in literature, such as DROC for German, and Litbank \citep{litbank} and FantasyCoref \citep{fantasy} for English. For non-literary texts, significant datasets include OntoNotes \citep{ontonotes} in English, and the ETRI\footnote{https://aiopen.etri.re.kr/corpusModel} and NIKL \citep{nikl} Coreference datasets for Korean (refer to Table ~\ref{tab:stats}).

% droc
\textbf{DROC} annotates coreference relations within 90 German novels, thereby pioneering the creation of coreference datasets specifically focused on characters within novels. It is recognized as the earliest dataset in this domain available to the public.

% litbank
\textbf{Litbank} annotates coreference relations across 210,532 tokens extracted from selected chapters of 100 notable works of English literature. Its annotations encompass ACE categories including people, locations, organizations, facilities, geopolitical entities, and vehicles \citep{ace}. Litbank establishes the first comprehensive annotation guidelines specifically tailored to coreferences within this domain.

%fantasycoref
\textbf{FantasyCoref} annotates coreference relations in a corpus of 367,891 tokens, sourced from 211 Grimm's Fairy Tales and three fantasy novels. It expands upon Litbank by covering all entity types in its annotations.

%ontonotes
\textbf{OntoNotes} serves as a benchmark in coreference resolution, with most recent models evaluating their performance on this dataset (\citealp{e2e}; \citealp{joshibert}; \citealp{spanbert}; \citealp{s2e}; \citealp{lingmess}). However, as highlighted by \citet{litbank}, OntoNotes spans non-literary domains, including news, conversations, and web contents.

%korean
In the domain of Korean coreference resolution, two pivotal resources are distinguished: \textbf{the ETRI dataset} and \textbf{the NIKL dataset}. The former comprises documents sourced from Wikipedia QA data. Conversely, the NIKL dataset integrates news and spoken language materials, such as broadcasts, lectures, and drama scripts.

\subsection{Literary-specific Considerations}
\citet{rosiger}, \citet{litbank}, and \citet{fantasy} have systematically identified and addressed specialized considerations relevant to coreference resolution within literary texts. These considerations can be broadly categorized into \textit{Changes in Entity} and \textit{Asymmetry in Knowledge}.\\

\noindent{\textbf{Changes in Entity}}
Classical coreference resolution models, as \citet{litbank} note, typically assess whether two mentions denote the same real-world entity. In the context of literary texts, however, challenges arise when (1) mentions do not correspond to tangible real-world entities, (2) entities undergo identity transformations over the course of a narrative, or (3) the same entity is depicted differently by the narrator. \citet{rosiger} explore these issues under the theme of Entity Development, further noting (4) the complexities introduced by the formation of groups by multiple entities and the resultant multi-entity references (refer to Section~\ref{sec:multi}).

\textit{Litbank} and \textit{FantasyCoref} address the first challenge by determining entity identity within the narrative world. However, their strategies diverge when dealing with the second and third scenarios. \textit{FantasyCoref}, for instance, posits that an entity undergoing any form of transformation within the narrative world, such as a lion turning into a white dove, retains its identity. \textit{Litbank}, in contrast, uses the example of France before and after the Revolution to suggest that identity depends on the narrative's emphasis on entity differences. Drawing from \citet{nearidentity}, \textit{Litbank} differentiates between refocusing, which accentuates differences, and neutralization, which underscores identity.

\textit{KoCoNovel} encounters similar challenges, where mentions emphasize differences despite a shared narrative world identity. We align with FantasyCoref's guidelines to simplify annotation. For instance, as demonstrated in example (1), we treat phrases like "that sweet, innocent, and cheerful girl" and "this vile woman" as referring to the same entity, despite their contrasting descriptions. 

\begin{quote}
(1) [Jeonghee]$_{x}$, sitting in the lonely tram heading to the boarding house church and thinking of [A]$_{y}$, detested the "era" that had transformed \textbf{[that sweet, innocent, and cheerful girl]$_{y}$} into \textbf{[this vile woman]$_{y}$}.
(Kim, \textit{Jeong-hee})
\end{quote}

\noindent{\textbf{Asymmetry in Knowledge}}
is identified as another significant challenge in literary coreference resolution (\citealp{rosiger}; \citealp{litbank}; \citealp{fantasy}). This issue arises from the different worlds constructed by the unique information sets accessible to readers, narrators, and characters, leading to varied annotation approaches based on the selected narrative perspective. While both \textit{Litbank} and \textit{FantasyCoref} base their annotations on \textit{the omniscient writer’s point of view}, \textit{KoCoNovel} offers annotations from the perspectives of both the omniscient author and readers. As shown in Table~\ref{tab:sample_b}, it becomes apparent that "Taehoon" and "Kyungsuk" are parts of a one-man show by "Ms. B" in the novel's climax. Accordingly, in the \textit{Reader} version, they are annotated as separate entities, but as one in the \textit{Omniscient} version.
\label{sec:asym}

\begin{table}[h!]
\centering
\resizebox{7.8cm}{!}{
\begin{tabular}{ll}
\toprule
\multicolumn{2}{c}{\textbf{Episodes}}\\
\midrule
\multirow{4}{*}{A} & "Oh, \textbf{\{Taehoon\}}! Would that really be okay?"\\
& It was the melodious voice of a woman.\\
& "If \textbf{\{Kyungsuk\}} likes it, how happy \textbf{\{I\}} would be."\\
& \textbf{\{The man\}}'s voice was clear, filled with passion.\\
\midrule
\multirow{7}{*}{B} & What a weird scene! The lights were still on, \\
& but the bed was covered with what looked like \\
& \textit{love letters} sent to students, all scattered around.\\
& In the middle of this mess, \textbf{\{Ms. B\}} sat up alone. ... \\
& With a face that looked kind of funny and\\
& desperate, \textbf{\{she\}} seemed to be waiting for a kiss. \\
& At the same time, \textbf{\{she\}} tried to sound like a man. \\
\end{tabular}}
\resizebox{7.8cm}{!}{
\begin{tabular}{cl}
\toprule
\multicolumn{2}{c}{\textbf{Coreference}}\\
\midrule
\multirow{2}{*}{\textbf{Reader}} &  [Taehoon, I, The man], [Kyungsuk], [Ms. B, \\
& she, she]\\
\midrule
\multirow{2}{*}{\textbf{Omniscient}} & [Taehoon, I, The man, Kyungsuk, Ms. B, \\
& she, she]\\
\bottomrule
\end{tabular}}
\caption{Examples of coreference from the perspectives of the omniscient author and readers in KoCoNovel. The episodes are from \textit{Ms. B and Love Letters} by Hyeon}
\label{tab:sample_b}
\end{table}

% criteria
\begin{table*}[t!]
\centering
\resizebox{\textwidth}{!}{%
\begin{tabular}{lcl}
\toprule
\textbf{Case} & \textbf{Character} & \textbf{Basis for Judgment} \\
\midrule
A goblin asking for a favor & Yes & Engagement in dialogue\\
\midrule
A cow crying at the protagonist's words & Yes & Emotional exchange with the protagonist\\
\midrule
A bird the protagonist believes is crying for him 
& No & Simple projection of the protagonist's emotions\\
\midrule
A ghost that the protagonist mistakenly sees & Maybe & Depends on what the entity is revealed to be
\\
\bottomrule
\end{tabular}}
\caption{Criteria for determining whether non-human entities in narratives are considered characters.}
\label{tab:character}
\end{table*}

\section{Annotation Process}
%collect-first
\noindent{\textbf{Preprocessing}} In this paper, we select 50 works from the Korean modern and contemporary novels that are in the public domain on Wikisource\footnote{https://ko.wikisource.org/wiki}, considering the diversity of writing styles (see Appendix ~\ref{sec:works}). The preprocessing involved correcting typos and incorrect line breaks within the text, and adjusting the spelling to match modern Korean grammar to enhance the dataset's universality and convenience for annotators. Spelling correction was conducted with the assistance of the Pusan National University Korean grammar checker\footnote{http://speller.cs.pusan.ac.kr/} and manual verification.\\

\begin{figure}[h!]
\centering
\includegraphics[width=7.8cm]{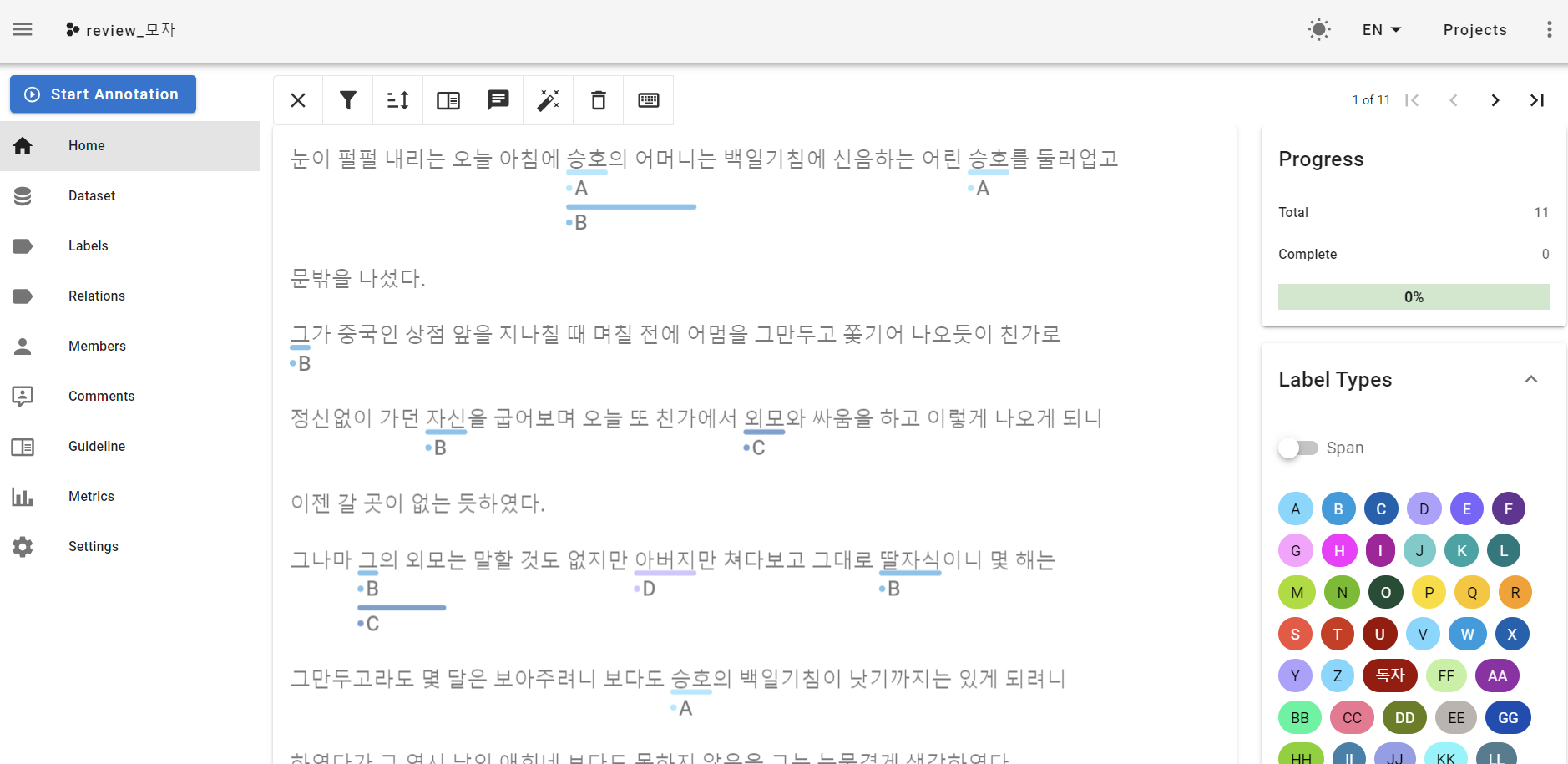}
\caption{Doccano Interface used for annotating coreference relations in Korean novels. Labels could be assigned arbitrarily, but consistency within each group should be maintained.}
\label{fig:doc}
\end{figure}

%collect-second 
\noindent{\textbf{Annotation}} The annotation process engaged 19 participants, including two authors, with three workers assigned to each literary work. They were tasked with identifying character mentions within the works, marking mentions that refer to the same entity with the same label. For direct quotes, they were instructed to mark a distinct speaker id to differentiate from the narrator. As represented in Figure~\ref{fig:doc}, Doccano program \citep{doccano} was used for the annotation process. Given that annotators reviewed coreference relations sequentially throughout the narrative, they were instructed to record any changes in coreference relations resulting from developments within the story. In the post-processing phase, the dataset was divided into two versions to reflect the perspectives of both the omniscient author and the readers.

\section{Annotation Guidelines}
The guidelines for the KoCoNovel Dataset are based on those of the ETRI and NIKL Corpus, reflecting the linguistic characteristics of Korean. They are tailored to address the nuances of Korean and the distinct aspects of novels. Details are provided below. For guidelines incorporated from other corpora, refer to Appendix~\ref{sec:other}.

\subsection{Definition of Character}
\label{sec:defchar}
In KoCoNovel dataset, coreference targets are mentions indicating characters, who are predominantly human but may occasionally be animals or inanimate objects. This study aims to provide clear definitions for scenarios that were previously considered as edge cases in DROC.\\

\noindent{\textbf{Human Characters}} Mentions of specific individuals are classified as character mentions, despite debates about whether single mentions in dialogue qualify as characters. Given the complexities of narrative structures, classifying entities based solely on \textit{Appearance} is deemed impractical. Therefore, annotators identified all mentions related to individuals, excluding idiomatic references to academics, religion (e.g.,\textit{ Confucius, Mencius, Jesus}), or time periods (e.g., \textit{the fifteenth year of King Daemusin of Goguryeo}). Additionally, groups, such as "the crowd" and "a group of boys" are treated as characters unless they are recognized as independent legal entities, such as "the magazine headquarters".\\

\noindent{\textbf{Non-Human Characters}} For non-human entities, their qualification as characters is based on whether they engage in dialogue or emotional interactions with other characters. The specific examples for this are outlined in Table~\ref{tab:character}.\\

\noindent{\textbf{Readers}} In certain narratives, narrators explicitly engage with readers. To accommodate this unique interaction, these instances are annotated using a distinct \textit{Reader} label, thereby introducing a novel element absent from earlier literary text-based coreference resolution datasets.

\begin{quote}
(2) "Wise \textbf{[reader]}! Why hesitate? Surely this profound and critical issue awaits \textbf{[your]} awareness, wisdom, and strength?" (Lee, \textit{City and Ghost})
\end{quote}

\subsection{Maximal Span} 
\label{sec:maxspan}
We adhered to the Maximal Span principle, which includes all modifiers of a noun, and revised its exceptions. Notably, the existing exceptions in the ETRI and NIKL Corpus are as follows:

\begin{itemize}
\item For [Name+Title] combinations, additionally extract the embedded name.
\item For pronouns and phrases that start with demonstrative adjectives, exclude prior modifiers from the span. 
\end{itemize}

We eliminated the redundant requirement to re-extract embedded names (condition 1) and introduced an additional condition: Exclude modifying clauses for proper nouns or noun phrases acting as proper nouns. This aims to reduce mentions exceeding 15 tokens and to align with Korean syntax.

\subsection{Generic vs Specific Mention}
\label{sec:versus}
Noun phrases can refer to general classes (generic mentions) or specific entities (specific mentions). In the context of noun phrases, specific mentions are identified through the use of grammatical markers (refer to Appendix ~\ref{sec:generic}). However, the Korean language presents unique challenges in this regard due to the absence of determiners, the lack of specific markers for proper nouns, and the common omission of the plural suffix \textquoteleft-tul\textquoteright. Moreover, because of its cultural context, characters are frequently depicted only through common nouns and noun phrases. Consequently, this study advances the guidelines established by NIKL, proposing refined principles for the identification and classification of generic and specific mentions within texts.

\begin{itemize}
\item Common nouns without modifiers are treated as specific mentions if consistently referring to a specific entity in a novel, and considered equivalent to proper names when no other entity could be referenced.

\item Noun phrases limited by demonstrative adjectives or expressions of exclusivity are also categorized as specific mentions (e.g., "a son like no other in the world").

\item Groups, undefined in number but identifiable by context (e.g., people in a square, farmers on the protagonist's land), are treated as specific mentions.

\end{itemize}

\subsection{Pronouns: Wuli, Cehuy}
In the Korean language, the first-person plural pronouns \textit{Wuli} and \textit{Cehuy} exhibit unique usage patterns. As a result, we excludes them from the targets for coreference annotation. They primarily distinguish in-groups from out-groups, with in-group interpretation varying by context \citep{wuli}. For instance, \textit{Wuli} in \textit{"Wuli Seungho"} (the grandson's name) during a conversation between a mother-in-law and a daughter-in-law may mean "my" or "our", depending on their psychological relationship. To mitigate potential confusion among annotators, we exclude these two pronouns.

\subsection{Nested Mentions in Proper Names}
Proper names are typically treated as atomic entities, with internal mentions unmarked. However, due to the nuances of culture of address terms in Korea, we introduce an exception. In Korea, it is common to refer to mothers by a format that combines their first child’s name with "emma" (meaning "Mom"), as in "Bobu emma"  for the mother of Bobu, effectively substituting their personal names in social interactions. Prior research on Korean coreference resolution has not provided guidelines for addressing such expressions. Consequently, this study categorizes this kind of expressions as proper names while permitting the annotation of the children's names they contain.

\subsection{Multiple Entities}
\label{sec:multi}
Numerous coreference resolution studies have traditionally distinguished between plural and singular mentions, annotating them as distinct entities. Conversely, \citet{zhou} annotates plural mentions with overlapping entities, thereby recognizing the connections between singular and plural mentions. In this paper, we provide distinct versions that both \textit{separate} and \textit{overlap} entities on plural mentions, accommodating previous models and addressing the lexical nuances of Korean (refer to Table~\ref{tab:multi_entity}).

In Korean, relationships such as \textit{mother and son} are encapsulated in single compound words (e.g., "moca"), presenting challenges for clarity within datasets. Consequently, while maintaining the predominant approach (\textit{Seperate}), we also permit the annotation of plural mentions with overlapping entities from singular mentions, provided the relationship between these entities is clearly defined (\textit{Overlapped}).

\begin{table}[h!]
\centering
\resizebox{7.8cm}{!}{
\begin{tabular}{cl}
\toprule
\multicolumn{2}{c}{Sentences}\\
\midrule
\textit{KO} & \textbf{\{나\}}는 \textbf{\{승호\}}를 등에 업었다.\\
\textit{Yale} & \textbf{\{na\}}nun \textbf{\{sungho\}}lul tungey epessta.\\
\textit{EN} & \textbf{\{I\}} carried \textbf{\{Seungho\}} on my back.\\
\midrule
\textit{KO} & \textbf{\{우리 모자\}}는 눈 내리는 거리를 방황했다.\\
\textit{Yale} & \textbf{\{wuli moca\}}nun nwun naylinun kelilul \\
& panghwanghayssta. \\
\textit{EN} & \textbf{\{The mother and son\}} wandered the snowy streets. \\ 
\end{tabular}}
\resizebox{7.8cm}{!}{
\begin{tabular}{ll}
\toprule
\multicolumn{2}{c}{Coreference} \\
\midrule
\textbf{Separate} & [I], [Seungho], [The mother and son]\\
\midrule
\multirow{2}{*}{\textbf{Overlapped}} & [I, The mother and son], [Seungho, The mother\\
& and son]\\ 
\bottomrule
\end{tabular}}
\caption{Example of multiple entities: The \textit{Separate} approach is utilized in OntoNotes, FantasyCoref, and many other datasets. \textit{Overlapped} is used in \citet{zhou}. \textit{Yale} means Yale Romanization}
\label{tab:multi_entity}
\end{table}

% chapter four
\section{Analysis}
\subsection{Inter-Annotator Agreement}
The annotation outcomes demonstrate a significant level of agreement among three annotators, as elaborated in Table~\ref{tab:iaa}. For the assessment of inter-annotator agreement (IAA), we utilize the coreference evaluation metrics employed by FantasyCoref—MUC \citep{muc}, B$^3$ \citep{b3}, and CEAF$_{\phi4}$ \citep{ceaf}. Additionally, we include LEA \citep{lea} for a more comprehensive analysis. 
Litbank achieved a MUC score of 95.5 for IAA based on a 10\% sample of its corpus, whereas KoCoNovel recorded an average MUC score of 94.53 across the entirety of its dataset. The attainment of such high IAA levels, even among non-expert annotators, can be attributed to the implementation of detailed guidelines, a recruitment screening test, and sustained communication with the research team.

% iaa
\begin{table}[h!]
\centering
\begin{tabular}{ccccc}
\toprule
&\textbf{MUC} & \textbf{B$^3$} & \textbf{CEAF$_{\phi4}$} & \textbf{LEA}\\
\midrule
Avg. & 94.53 & 86.85 & 79.71 & 86.54\\
\bottomrule
\end{tabular}
\caption{Inter-annotator agreement scores measured
by the evaluation metrics. The scores are based on the F1 scores, averaged from pairwise calculations among three annotators.}
\label{tab:iaa}
\end{table}

\subsection{Statistics}
Table~\ref{tab:basic} presents the descriptive statistics for KoCoNovel. Based on 50 full-length Korean short stories, KoCoNovel annotates coreference resolution for a total of 178,957 tokens. Regarding the treatment of multiple entities, the \textit{Separate} approach encompasses 1428 entities, whereas the \textit{Overlapped} method accounts for 1027 entities. The entity count per narrative, representing characters, varies significantly, ranging from a maximum of 97 to a minimum of 5. Moreover, there are 19,030 mentions in total, with individual stories containing up to 2,241 mentions and at least 103 mentions. These statistics underscore the diversity and complexity of character representation within KoCoNovel, highlighting the dataset's potential for in-depth coreference resolution studies in the context of Korean narrative literature.

\begin{table}[h!]
\centering
\resizebox{7.8cm}{!}{
\begin{tabular}{llcccc}
\toprule
& & \multirow{2}{*}{\textbf{Total}} &  \multicolumn{3}{c}{\textbf{{\# per Document}}}\\ \cmidrule(lr){4-6}
&&& \textbf{Avg.} & \textbf{Max.} & \textbf{Min.}\\
\midrule
Sentences & & 17975 & 359.5 & 1950 & 107\\
Tokens & & 178928 & 3578.6 & 19875 & 1087\\
Mentions & & 19030 & 380.6 & 2241 & 103\\
\midrule
\multirow{2}{*}{Entities} & \textit{Separate}& 1418 & 28.36 & 126 & 6\\
& \textit{Overlapped} & 1027 & 20.54 & 97 & 5\\
\bottomrule
\end{tabular}}
\caption{Basic statistics of KoCoNovel: Statistics are based on the omniscient author's point of view, as the distinction between reader's and omniscient views appears only in select works. (refer to Appendix~\ref{sec:omni})}
\label{tab:basic}
\end{table}

\subsection{Comparison with NIKL Dataset}
We compare KoCoNovel with the NIKL Corpus, the only fully public and largest coreference resolution dataset in Korean. The NIKL Corpus includes both written corpora derived from news sources (\textit{News}) and spoken corpora comprising broadcast scripts and dialogues (\textit{Spoken}). We also calculate \textit{Antecedent Distance}, the average interval between mentions of a single entity, and \textit{Spread}, the distance between an entity's first and last mentions.

The analysis indicates that KoCoNovel consists of documents that are generally lengthier than those in the NIKL Corpus. Accordingly, the Spread within KoCoNovel documents is broader compared to that observed within NIKL. However, KoCoNovel's Antecedent Distance falls between the lengths noted in NIKL’s spoken and written texts. Direct comparisons present challenges due to NIKL's comprehensive coverage of entities versus KoCoNovel's focus on character mentions. Nevertheless, despite its focused scope, KoCoNovel encompasses a relatively greater number of entities.

\begin{table}[h!]
\centering
\resizebox{7.8cm}{!}{
\begin{tabular}{ccccc}
\toprule
& \multirow{2}{*}{\textbf{\makecell{\# per\\Document}}} & \multirow{2}{*}{\textbf{KoCoNovel}} & \multicolumn{2}{c}{\textbf{NIKL}}\\
&&&\textbf{News} & \textbf{Spoken}\\
\midrule
\multirow{3}{*}{Tokens} & Avg. & 3578.5 & 275.3 & 2379.3 \\
& Max. & 19875 & 2347 & 15510 \\
& Min. & 1087 & 41 & 147 \\
\midrule
\multirow{3}{*}{Entities} & Avg.&
28.36 & 12.94 & 39.78\\
& Max. & 126 & 60 & 294 \\
& Min. & 6 & 1 & 1 \\
\midrule
\multirow{3}{*}{\makecell{Antecedent\\Distance}} & Avg.& 70.7 & 38.21 & 163.22 \\
& Max. & 10279 & 1863 & 15406 \\
& Min. & 1 & 1 & 1 \\
\midrule
\multirow{3}{*}{Spread} & Avg. &
1583.3 & 119.5 & 847.6 \\
& Max. & 19603 & 2318 & 15488 \\
& Min. & 1 & 1 & 1 \\
\bottomrule
\end{tabular}}
\caption{Comparison between KoCoNovel and NIKL. Regarding \textit{Spread} and \textit{Antecedent Distance}, the distance is not calculated for nested mentions.}
\label{tab:compare}
\end{table}

\subsection{Categories of Mentions}
We analyze KoCoNovel's 19,030 mentions by first categorizing them into pronominals, nouns, and noun phrases, and then classifying them by Korean address term types as outlined by \citet{kang}. Findings are presented in Table~\ref{tab:kinship}.

\begin{table}[b!]
\centering
\resizebox{7.8cm}{!}{
\begin{tabular}{llll}
\toprule
\textbf{Cat.} & \textbf{Count} & \textbf{Subcat.} & \textbf{Count}\\
\midrule
\textbf{Pronominal} & 5846 (30.7\%) & - & - \\
\textbf{Proper Name} & 4338 (22.8\%) & - & -\\
\midrule
\multirow{3}{*}{\textbf{Single Noun}} & \multirow{3}{*}{\textbf{4591 (24.1\%)}} & Kinship & \textbf{1742 (9.2\%)}\\
& & Titles & 581 (3.1\%)\\
& & etc & 2268 (11.8\%)\\
\midrule
\multirow{3}{*}{\textbf{Noun Phrase}} & \multirow{3}{*}{4255 (22.4\%)} & Name+$\alpha$ & 874 (4.6\%)\\
& & Kinship & \textbf{845 (4.4\%)}\\
& & etc & 2536 (13.3\%)\\
\midrule
\textbf{Total} & 19030 (100\%) & - & - \\
\bottomrule
\end{tabular}}
\caption{Classification of Mentions in KoCoNovel}
\label{tab:kinship}
\end{table}

First, nouns and noun phrases related to kinship represent 1,742 and 845 cases, respectively, constituting 13.6\% of the overall count. This observation not only indicates the prevalence of familial themes within the KoCoNovel corpus but also mirrors the Korean cultural convention of employing kinship terms for individuals who are not relatives, a practice known as \textit{Fictive Kinship}. The predominance of nouns over noun phrases can be attributed to the linguistic tendency in Korean to frequently omit possessive markers such as "my".

Second, the analysis of KoCoNovel reveals a significant prevalence of single nouns devoid of modifiers, accounting for 24\% of the total. This pattern suggests the lack of a grammatical distinction between common nouns and specific nouns in Korean, attributable to the non-existence of determiner. Moreover, a majority of these unmodified common nouns are associated with kinship or titles, underscoring their utilization as nominal substitutes.

\section{Experiment Setup}
We aim to assess how significantly our corpus specialized for literary works, improves coreference resolution in literary texts compared to existing non-literary corpora. We partition the KoCoNovel dataset into training, development, and testing segments, ensuring that there are no overlapping novels between them. We employ a BERT-based coreference resolution model \citep{joshibert}, training it solely on this partitioned KoCoNovel dataset. As a baseline, we train models exclusively on the NIKL dataset and a combined dataset of NIKL and KoCoNovel training data. Details on the dataset division and the models utilized are presented below.\\

\begin{table}[b!]
\centering
\resizebox{7.8cm}{!}{
\begin{tabular}{lcccccc}
\toprule
& \multicolumn{2}{c}{\textbf{Split1}} & \multicolumn{2}{c}{\textbf{Split2}} & \multicolumn{2}{c}{\textbf{Split3}} \\
\cmidrule(l){2-3} \cmidrule(lr){4-5} \cmidrule(r){6-7}
& Episodes & Works & Episodes & Works & Episodes & Works\\
\midrule
Train  & 782  & 45  & 799 & 44 & 784  & 44\\
Dev    & 52 & 2  & 31 & 3  & 45 & 3 \\
Test   & 39 & 3 & 43 & 3 & 44 & 3\\
\midrule
Total  & 873 & 50 & 873 & 50 & 873 & 50\\ 
\bottomrule
\end{tabular}}
\caption{Episodes and Works in Splits. For the IDs of the works in the development and test sets in each split, refer to Appendix~\ref{sec:split_id}}
\label{my-label}
\end{table}

\begin{table*}[ht]
\centering
\resizebox{\textwidth}{!}{
\begin{tabular}{lcccccccccccc}
\toprule
& \multicolumn{12}{c}{split 1}\\
\cmidrule{2-13}
& \multicolumn{3}{c}{MUC} & \multicolumn{3}{c}{B$^3$} & \multicolumn{3}{c}{CEAF$_{\phi4}$} &\\
\cmidrule(l){2-4} \cmidrule(l){5-7} \cmidrule(l){8-10}
& P & R & F1 & P & R & F1 & P & R & F1  & Avg.P & Avg.R & Avg.F1\\
\midrule
KoCoNovel (Ours) & 77.41 & 75.43 &  \textbf{76.40} &
60.07 & 62.38 & \textbf{61.20} &
\textbf{66.71} & \textbf{41.53} & \textbf{51.19} &
68.06 & 59.78 & 62.93 \\
+ AA & 74.52 & 73.38 & 73.95 & 
48.82 & 62.71 & 54.90 &
63.70 & 32.05 & 42.65 & 
62.35 & 56.05 & 57.16
\\
+ SC & \textbf{79.25} & 68.43 & 73.44 & 
\textbf{66.25} & 55.08 & \textbf{60.15} &
63.95 & 41.43 & 50.28 &
69.82 & 54.98 & 61.29\\
\midrule
NIKL & 54.31 & 72.01 & 61.92 & 
29.20  & 60.15 & 39.32 &
28.56 & 26.41 & 27.44 &
37.36 & 52.86 & 42.89\\
NIKL+KoCoNovel & 67.08 & \textbf{83.11} & 74.24 & 
42.54 & \textbf{72.81} & 53.70 &
54.81 & 34.82 & 42.58 &
54.81 & 63.58 & 56.84\\
\midrule
\end{tabular}}

\resizebox{\textwidth}{!}{
\begin{tabular}{lcccccccccccc}
& \multicolumn{12}{c}{split 2}\\
\cmidrule{2-13}
& \multicolumn{3}{c}{MUC} & \multicolumn{3}{c}{B$^3$} & \multicolumn{3}{c}{CEAF$_{\phi4}$} &\\
\cmidrule(l){2-4} \cmidrule(l){5-7} \cmidrule(l){8-10}
& P & R & F1 & P & R & F1 & P & R & F1  & Avg.P & Avg.R & Avg.F1\\
\midrule
KoCoNovel (Ours) & 72.25 & 57.89 & 64.28 & 
55.28 & 46.40 & 50.45 &
58.79 & 29.54 & 39.33 &
62.10 & 44.61 &  51.35\\
+ AA & 74.31 & 59.42 & 66.04 & 
55.70 & 50.14 & 52.78 & 
\textbf{62.65} & 29.54 & 40.15 &
64.22 & 46.37 & 52.99\\
+ SC & \textbf{74.32} & 55.52 & 63.56 &
\textbf{60.06} & 45.51 & 51.79 & 
60.24 & \textbf{31.84} & \textbf{41.66} &
64.88 & 44.29 & 52.33 \\
\midrule
NIKL & 54.31 & \textbf{72.01} & 61.92 & 
29.20 & \textbf{60.15} & 39.32 & 
28.56 & 26.41 & 27.44 & 
37.36 & 52.86 & 42.89 \\
NIKL+KoCoNovel & 70.78 & 66.21 & \textbf{68.42} &
50.86 & 57.24 & \textbf{53.86} & 
56.89 & \textbf{31.84} & 40.83 & 
59.51 & 51.76 & 54.37\\
\midrule
\end{tabular}}

\resizebox{\textwidth}{!}{
\begin{tabular}{lcccccccccccc}
& \multicolumn{12}{c}{split 3}\\
\cmidrule{2-13}
& \multicolumn{3}{c}{MUC} & \multicolumn{3}{c}{B$^3$} & \multicolumn{3}{c}{CEAF$_{\phi4}$} &\\
\cmidrule(l){2-4} \cmidrule(l){5-7} \cmidrule(l){8-10}
& P & R & F1 & P & R & F1 & P & R & F1  & Avg.P & Avg.R & Avg.F1\\
\midrule
KoCoNovel (Ours) & 79.29 & 66.77 & 72.49 &
62.72 & 58.15 & \textbf{60.35} & 
\textbf{68.15} & \textbf{30.42} & \textbf{42.06} & 
70.05 & 51.78 & 58.30\\
+ AA & \textbf{82.38} & 62.56 & 71.11 &
\textbf{67.23} & 51.70 & 58.45 & 
66.92 & 29.87 & 41.31 & 
72.18 & 48.04 & 56.96\\
+ SC  & 79.63 & 65.26 & 71.74 &
59.74 & 53.57 & 56.49 & 
65.57 & 27.58 & 38.83 &
68.32 & 48.80 & 55.68\\
\midrule
NIKL & 60.37 & 69.17 & 64.47 &
35.13 & 59.69 & 44.23 &
43.36 & 24.01 & 30.90 & 
46.29 & 50.96 & 46.53 \\
NIKL+KoCoNovel & 71.75 & \textbf{75.64} & \textbf{73.65} & 
46.36 & \textbf{66.90} & 54.77 &
59.89 & 26.73 & 36.96 & 
59.33 & 56.42 & 55.13\\
\bottomrule
\end{tabular}}
\caption{The highest performance results from the models we trained—using KoCoNovel, NIKL, and a combination of both datasets—were measured using the standard coreference evaluation metrics: MUC, B$^3$, and CEAF$_{\phi4}$ (Precision: P, Recall: R).}
\label{tab:performance}
\end{table*}

\noindent \textbf{Dataset Division}
We conduct experiments in cross-domain settings. First, we divide the 50 works into segments of approximately 20 sentences each, obtaining 873 episodes. We select 2-3 different works to compile the test and development sets, aiming for approximately 30-50 episodes each. We choose the \textit{Omniscient \& Separate} version of KoCoNovel and convert morpheme units to word units to adhere to a traditional approach.\\

\noindent \textbf{Models} We conduct experiments with the BERT-based coreference resolution model \citep{joshibert} with two types of higher-order inference (HOI) methods. We only utilize klueBERT-base \citep{klue} as a base model, given the absence of public Korean versions of SpanBERT \citep{spanbert} and Longformer \citep{longformer}. For HOI, we use attended antecedent (\textit{AA}; \citealp{leehoi}) and span clustering (\textit{SC}; \citealp{sc2020}). For processing long documents, we employ an independent split approach rather than an overlapping split \citep{joshibert}. For details on hyper-parameters, see Appendix ~\ref{sec:hyper}.

\section{Experiment Results} 
For each dataset split, we select the best-performing model on the development (dev) set for evaluation on the test set. The primary criterion for assessing performance on the dev set is generally the average F1 score. However, given the NIKL dataset encompasses all entities and not just characters, the average recall score serves as our benchmark for models involving the NIKL dataset.

Table~\ref{tab:performance} presents the best performances of models trained solely on KoCoNovel, those trained on a combined dataset of NIKL and KoCoNovel, and models trained exclusively on NIKL. Regarding HOI, no consistent enhancement is observed, and we report performances without HOI for models involving the NIKL dataset. The benefits of training on the KoCoNovel dataset are evident in Splits 1 and 3, where it outperforms the NIKL-only approach across all metrics. Nonetheless, in Split 2, the NIKL-only model demonstrated superior recall. Furthermore, the higher recall values for the model combining NIKL and KoCoNovel, compared to the KoCoNovel-only model, indicate there is room for enhancement. Given the significant size difference between the NIKL and KoCoNovel datasets, KoCoNovel has proven effective. However, there is a need for a larger dataset and a more refined approach to integrate NIKL and KoCoNovel effectively.

\section{Conclusion}
In this study, we present KoCoNovel, a novel dataset derived from Korean literary texts for character coreference, featuring 178,000 tokens from 50 modern and contemporary Korean novels. Notably, 24\% of character mentions in KoCoNovel are single common nouns, reflecting Korean's unique address terms that prioritize social roles over names. Upon evaluating a BERT-based coreference resolution model with KoCoNovel, we observed significant improvements in performance compared to the NIKL corpus. However, models trained on a combination of NIKL and KoCoNovel datasets often yielded the best performance, and overall, the performance varies depending on the dataset split. Based on these findings, we anticipate that further refinement of coreference models, specifically tailored to leverage the unique characteristics of KoCoNovel, will enable an even more nuanced understanding and processing of Korean literary text.

\section*{Acknowledgements}
This work was supported by the Undergraduate Independent Research Program of the Faculty of Liberal Education, Seoul National University.

\bibliography{act2emo}

\begin{thebibliography}{26}
\expandafter\ifx\csname natexlab\endcsname\relax\def\natexlab#1{#1}\fi

\bibitem[{Bagga and Baldwin(1998)}]{b3}
Amit Bagga and Breck Baldwin. 1998.
\newblock Algorithms for scoring coreference chains.
\newblock In \emph{The first international conference on language resources and evaluation workshop on linguistics coreference}, volume~1, pages 563--566. Citeseer.

\bibitem[{Bamman et~al.(2019)Bamman, Lewke, and Mansoor}]{litbank}
David Bamman, Olivia Lewke, and Anya Mansoor. 2019.
\newblock An annotated dataset of coreference in english literature.
\newblock \emph{arXiv preprint arXiv:1912.01140}.

\bibitem[{Beltagy et~al.(2020)Beltagy, Peters, and Cohan}]{longformer}
Iz~Beltagy, Matthew~E Peters, and Arman Cohan. 2020.
\newblock Longformer: The long-document transformer.
\newblock \emph{arXiv preprint arXiv:2004.05150}.

\bibitem[{Daudert(2020)}]{doccano}
Tobias Daudert. 2020.
\newblock A web-based collaborative annotation and consolidation tool.
\newblock In \emph{Proceedings of the Twelfth Language Resources and Evaluation Conference}, pages 7053--7059.

\bibitem[{Han et~al.(2021)Han, Seo, Kang, Kim, Choi, Song, and Choi}]{fantasy}
Sooyoun Han, Sumin Seo, Minji Kang, Jongin Kim, Nayoung Choi, Min Song, and Jinho~D Choi. 2021.
\newblock Fantasycoref: Coreference resolution on fantasy literature through omniscient writer’s point of view.
\newblock In \emph{Proceedings of the Fourth Workshop on Computational Models of Reference, Anaphora and Coreference}, pages 24--35.

\bibitem[{Hovy et~al.(2006)Hovy, Marcus, Palmer, Ramshaw, and Weischedel}]{ontonotes}
Eduard Hovy, Mitch Marcus, Martha Palmer, Lance Ramshaw, and Ralph Weischedel. 2006.
\newblock Ontonotes: the 90\% solution.
\newblock In \emph{Proceedings of the human language technology conference of the NAACL, Companion Volume: Short Papers}, pages 57--60.

\bibitem[{Joshi et~al.(2020)Joshi, Chen, Liu, Weld, Zettlemoyer, and Levy}]{spanbert}
Mandar Joshi, Danqi Chen, Yinhan Liu, Daniel~S Weld, Luke Zettlemoyer, and Omer Levy. 2020.
\newblock Spanbert: Improving pre-training by representing and predicting spans.
\newblock \emph{Transactions of the association for computational linguistics}, 8:64--77.

\bibitem[{Joshi et~al.(2019)Joshi, Levy, Weld, and Zettlemoyer}]{joshibert}
Mandar Joshi, Omer Levy, Daniel~S Weld, and Luke Zettlemoyer. 2019.
\newblock Bert for coreference resolution: Baselines and analysis.
\newblock \emph{arXiv preprint arXiv:1908.09091}.

\bibitem[{Kang(2005)}]{kang}
Hyunja Kang. 2005.
\newblock The charateristics of address terms in korean: From a sociolinguistic point of view.
\newblock \emph{The language and Culture}, 1(2):223--243.

\bibitem[{Kirstain et~al.(2021)Kirstain, Ram, and Levy}]{s2e}
Yuval Kirstain, Ori Ram, and Omer Levy. 2021.
\newblock Coreference resolution without span representations.
\newblock \emph{arXiv preprint arXiv:2101.00434}.

\bibitem[{Krug et~al.(2017)Krug, Weimer, Reger, Macharowsky, Feldhaus, Puppe, and Jannidis}]{droc}
Markus Krug, Lukas Weimer, Isabella Reger, Luisa Macharowsky, Stephan Feldhaus, Frank Puppe, and Fotis Jannidis. 2017.
\newblock Description of a corpus of character references in german novels-droc [deutsches roman corpus].

\bibitem[{Lee(2007)}]{wuli}
Han-gyu Lee. 2007.
\newblock A pragmatic and sociocultural approach to the so-called 1st person pronoun wuli in korean.
\newblock \emph{Discourse and Cognition}, 14(3):155--178.

\bibitem[{Lee et~al.(2017)Lee, He, Lewis, and Zettlemoyer}]{e2e}
Kenton Lee, Luheng He, Mike Lewis, and Luke Zettlemoyer. 2017.
\newblock End-to-end neural coreference resolution.
\newblock \emph{arXiv preprint arXiv:1707.07045}.

\bibitem[{Lee et~al.(2018)Lee, He, and Zettlemoyer}]{leehoi}
Kenton Lee, Luheng He, and Luke Zettlemoyer. 2018.
\newblock Higher-order coreference resolution with coarse-to-fine inference.
\newblock \emph{arXiv preprint arXiv:1804.05392}.

\bibitem[{Luo(2005)}]{ceaf}
Xiaoqiang Luo. 2005.
\newblock On coreference resolution performance metrics.
\newblock In \emph{Proceedings of human language technology conference and conference on empirical methods in natural language processing}, pages 25--32.

\bibitem[{Moosavi and Strube(2016)}]{lea}
Nafise~Sadat Moosavi and Michael Strube. 2016.
\newblock Which coreference evaluation metric do you trust? a proposal for a link-based entity aware metric.
\newblock In \emph{Proceedings of the 54th annual meeting of the association for computational linguistics}, volume~1, pages 632--642. Association for Computational Linguistics.

\bibitem[{NIKL(2021)}]{nikl}
National Institute of Korean~Language NIKL. 2021.
\newblock \href {https://kli.korean.go.kr/corpus} {Nikl korean coreference resolution corpus 2020}.
\newblock \url{https://kli.korean.go.kr/corpus}.

\bibitem[{Otmazgin et~al.(2022)Otmazgin, Cattan, and Goldberg}]{lingmess}
Shon Otmazgin, Arie Cattan, and Yoav Goldberg. 2022.
\newblock Lingmess: Linguistically informed multi expert scorers for coreference resolution.
\newblock \emph{arXiv preprint arXiv:2205.12644}.

\bibitem[{Park et~al.(2021)Park, Moon, Kim, Cho, Han, Park, Song, Kim, Song, Oh et~al.}]{klue}
Sungjoon Park, Jihyung Moon, Sungdong Kim, Won~Ik Cho, Jiyoon Han, Jangwon Park, Chisung Song, Junseong Kim, Yongsook Song, Taehwan Oh, et~al. 2021.
\newblock Klue: Korean language understanding evaluation.
\newblock \emph{arXiv preprint arXiv:2105.09680}.

\bibitem[{Recasens et~al.(2011)Recasens, Hovy, and Mart{\'\i}}]{nearidentity}
Marta Recasens, Eduard Hovy, and M~Ant{\`o}nia Mart{\'\i}. 2011.
\newblock Identity, non-identity, and near-identity: Addressing the complexity of coreference.
\newblock \emph{Lingua}, 121(6):1138--1152.

\bibitem[{R{\"o}siger et~al.(2018)R{\"o}siger, Schulz, and Reiter}]{rosiger}
Ina R{\"o}siger, Sarah Schulz, and Nils Reiter. 2018.
\newblock Towards coreference for literary text: Analyzing domain-specific phenomena.
\newblock In \emph{Proceedings of the Second Joint SIGHUM Workshop on Computational Linguistics for Cultural Heritage, Social Sciences, Humanities and Literature}, pages 129--138.

\bibitem[{Shanahan(1997)}]{culture}
Daniel Shanahan. 1997.
\newblock Articulating the relationship between language, literature, and culture: Toward a new agenda for foreign language teaching and research.
\newblock \emph{The Modern Language Journal}, 81(2):164--174.

\bibitem[{Vilain et~al.(1995)Vilain, Burger, Aberdeen, Connolly, and Hirschman}]{muc}
Marc Vilain, John~D Burger, John Aberdeen, Dennis Connolly, and Lynette Hirschman. 1995.
\newblock A model-theoretic coreference scoring scheme.
\newblock In \emph{Sixth Message Understanding Conference (MUC-6): Proceedings of a Conference Held in Columbia, Maryland, November 6-8, 1995}.

\bibitem[{Walker et~al.(2006)Walker, Strassel, Medero, and Maeda}]{ace}
Christopher Walker, Stephanie Strassel, Julie Medero, and Kazuaki Maeda. 2006.
\newblock Ace 2005 multilingual training corpus.
\newblock \emph{Linguistic Data Consortium, Philadelphia}, 57:45.

\bibitem[{Xu and Choi(2020)}]{sc2020}
Liyan Xu and Jinho~D Choi. 2020.
\newblock Revealing the myth of higher-order inference in coreference resolution.
\newblock \emph{arXiv preprint arXiv:2009.12013}.

\bibitem[{Zhou and Choi(2018)}]{zhou}
Ethan Zhou and Jinho~D Choi. 2018.
\newblock They exist! introducing plural mentions to coreference resolution and entity linking.
\newblock In \emph{Proceedings of the 27th International Conference on Computational Linguistics}, pages 24--34.

\end{thebibliography}

\appendix

\section{Guidelines Derived from Existing Datasets}
\label{sec:other}

\noindent\textbf{Span Unit} Following NIKL guidelines, we adopt morpheme units instead of word units for spans, aligning with the intuition of Korean speakers. However, we do not annotate morphemes at units smaller than a syllable.\\

\noindent\textbf{Singletons} A singleton is a mention that does not have an anaphoric relationship. We include singletons following the criteria from Litbank and to facilitate Character Detection.\\

\noindent\textbf{Copulae and Apposition} Copulae and apposition are considered unique structures in coreference resolution, observed in both Korean and English. However, we do not label copulae and apposition with separate tags, adhering to the standards set by ETRI and NIKL.

\section{Training Details}
\label{sec:hyper}
We utilized the codes from the Emory Language and Information Toolkit (ELIT) as outlined by \citet{sc2020}. Documents are split into independent segments with a maximum of 128 word-pieces. Models trained exclusively on KoCoNovel underwent 24 epochs of training, whereas other models were trained for 5 epochs. For additional hyperparameters, including those related to BERT and specific task parameters, we adhered to the guidelines set forth by \citet{sc2020}.

\begin{table}[h!]
\section{The Distinction between Reader’s and Omniscient Views}
\label{sec:omni}
\centering
\resizebox{7.8cm}{!}{
\begin{tabular}{cl}
\toprule
\textbf{Work ID} & \textbf{Contents} \\
\midrule
\multirow{3}{*}{0} & It is revealed that the "male lover" and \\ 
& "female lover" are actually part of a \\
& one-man show by "Ms. B".\\
\midrule
\multirow{2}{*}{13} & It is revealed that the "thief who stole the rice" \\ 
& is "Eung-o" (the owner of the rice field). \\
\midrule
\multirow{3}{*}{16} & It is revealed that the "God" the protagonist \\
& believes in and the "Heavenly judge" the \\
& protagonist meets after death are the same. \\
\midrule
\multirow{2}{*}{36} & It is revealed that the protagonist's "father" \\
& is "Director Jo". \\
\midrule
\multirow{2}{*}{49} & The culprit of three different incidents is \\
& revealed to be "Hyunam's father".\\
\bottomrule
\end{tabular}}
\caption{Works Illustrating the Distinction Between Reader's and Omniscient Viewpoints: For titles corresponding to Work IDs, refer to Appendix~\ref{sec:works}.}
\end{table}

\begin{table}[h!]
\section{Works in Development
and Test sets}
\label{sec:split_id}
\centering
%\resizebox{7.8cm}{!}{
\begin{tabular}{cll}
\toprule
Split & Dev & Test \\
\midrule
\textbf{1} & [13, 14] & [30, 31] \\
\midrule
\textbf{2} & [47, 48, 49] & [9, 10, 11]\\
\midrule
\textbf{3} & [2, 3, 4] & [25, 26, 27]\\
\bottomrule
\end{tabular}%}
\caption{The IDs of Works in the Development
and Test sets for Each Split: For titles corresponding to Work IDs, refer to Appendix~\ref{sec:works}.}
\end{table}

\begin{table*}[t!]
\section{Distinction Between Generic and Specific Mentions in English}
\label{sec:generic}
\centering
\resizebox{\textwidth}{!}{
\begin{tabular}{lcccccc}
\toprule
& \multicolumn{2}{c}{\textbf{<< More Specific}} && \multicolumn{2}{c}{\textbf{Less Specific >>}}\\
\cmidrule{2-6}
& \textbf{Proper Name} 	& \textbf{Pronominal} & \textbf{Definitive NP} & \textbf{Indefinite NP} & \textbf{Non-specific NP}\\
\midrule
Example &John & he& the linguist&a linguist I know& noted linguist\\
\midrule
Specific mention to Annotate & Yes & Yes & Yes & No & No \\
\midrule
Distinguishability in Korean & No & Yes & No & No & No\\
\bottomrule
\end{tabular}}
\caption{Examples of general and specific mentions in OntoNotes guidelines. \textit{Distinguishability} means whether this category can be distinguished by grammatical markers in Korean.}
\end{table*}

\begin{table*}[h!]
\section{The list of Korean novels}
\label{sec:works}
\centering
\resizebox{\textwidth}{!}{
\begin{tabular}{ccccc}
\toprule
\textbf{ID} & \textbf{Korean Title} & \textbf{Translated Title} & \textbf{Yale Romanization of Title} & \textbf{Romanization of Author}\\
\midrule
1 & B사감과 러브레터	
& Ms. B and Love Letters 
& B sakamkwa lepuleythe	
& Hyeon Jingeon\\

2 & 가애자 
& A Sweetheart 
& kaayca 
& Kim Namcheon\\

3& 가상의 불량소녀 
& A Fictional Rogue Girl 
& kasanguy pwullyangsonye
& Lee Iksang \\

4 & 결혼식 
& Wedding 
& kyelhonsik 
& Kim Dongin \\

5& 경희 
& Kyung-hee 
& kyenghuy & Na Hyesuk \\

6 & 괴승신수 
& A Strange Monk & koysungsinswu
& Yun Baeknam \\

7 & 그 여자 & The Woman 
& ku yeca & Kang Gyeongae \\

8 & 그믐달 & Dark Moon 
& kumumtal & Lee Iksang \\

9 & 금 따는 콩밭 
& A Field of Beans Where Gold is Buried 
& kum ttanun khongpath & Kim Yujeong \\

10 & 노다지 & Rich Mine & Nodachi & Kim Yujeong \\
11 & 대동강은 속삭인다 & The Daedong River Whispers & Daedongkang-ŭn Soksag'inda & Kim Dongin \\
12 & 도시와 유령 & City and Ghost & Dosŭiwa Yuryŏng & Yi Hyoseok \\
13 & 따라지 & Ddaraji & Ttaraji & Kim Yujeong \\
14 & 만무방 & Manmubang & Manmubang & Kim Yujeong \\
15 & 먼동이 틀 때 & When the Sun Rises & Mŏndong-i T'ŭl Ttae & Choe Seohae \\
16 & 메밀꽃 필 무렵 & When Buckwheat Flowers Blossom & Memilkkot Pil Muryŏp & Yi Hyoseok \\
17 & 명문 & Good Sentences & Myŏngmun & Kim Dongin \\
18 & 모르는 여인 & The Unknown Woman & Moreunŭn Yŏin & Lee Gwangsu \\
19 & 모자 & The Mother and the Son & Moja & Kang Gyeongae \\
20 & 무명초 & Anemone & Mumyŏngcho & Choe Seohae \\
21 & 박첨지의 죽음 & The Death of Park Chum-ji & Pakch'ŏmji Ŭi Jugŭm & Kim Dongin \\
22 & 벙어리 삼룡이 & The Mute Samurai & Pongŏri Samryong-i & Na Dohyang \\
23 & 봄봄 & Spring, Spring & Pombom & Kim Yujeong \\
24 & 부자 & The Father and the Son & Buja & Kang Gyeongae \\
25 & 불 & Fire & Pul & Hyeon Jingeon \\
26 & 빈처 & A Poverty-Stricken Wife & Pinch'o & Hyeon Jingeon \\
27 & 사위 & The Son-in-Law & Sawi & Lee Mu-yeong \\
28 & 소년의 비애 & A Boy's Woes & Sonyŏn-ŭi Pi-ae & Lee Gwangsu \\
29 & 송동이 & The Calf & Songdong'i & Kim Dongin \\
30 & 순정의 호동왕자 & Purehearted Prince Hodong & Sunch'ŏng-ŭi Hodongwangja & Yun Baengnam \\
31 & 술 권하는 사회 & Society Authorizing Alcohol & Sul Kwŏnhanŭn Sahoe & Hyeon Jingeon \\
32 & 신문지와 철창 & The Newspaper and the Cage & Sinmunjiwa Ch'ŏlchang & Hyeon Jingeon \\
33 & 우연의 기적 & Miracles of Coincidence & Uyŏn-ŭi Kijŏk & Yun Baengnam \\
34 & 운수 좋은 날 & A Lucky Day & Unsu Chohŭn Nal & Hyeon Jingeon \\
35 & 원수로 은인 & Enemies to Benefactors & Wŏnsuro Ŭnin & Yun Baengnam \\
36 & 유무 & Presence & Yumu & Kang Gyeongae \\
37 & 의심의 소녀 & Girl of Doubt & Ŭisim-ŭi Sonyŏ & Kim Myeongsun \\
38 & 이식과 도승 & Eisig and Doseung & Isikkwa Tosŭng & Yun Baeng \\
39 & 장미 병들다 & The Rose is Diseased & Changmi Pyŏngdŭlda & Yi Hyoseok \\
40 & 정열의 낙랑공주 & Passionate Princess Nangnang & Chŏngyŏl-ŭi Nangnanggongju & Kim Dongin \\
41 & 정희 & Jeong-hee & Chŏnghŭi & Na Dohyang \\
42 & 청춘 & Youth & Ch'ŏngch'un & Kim Yujeong \\
43 & 총각과 맹꽁이 & The Bachelor and the Blind Eel & Ch'onggakgwa Maengkkong'i & Choe Seohae \\
44 & 탈출기 & Escape Chronicles & Talch'ulgi & Kang Gyeongae \\
45 & 파금 & Break at Dawn & P'akŭm & Choe Seohae \\
46 & 해돋이 & Sunrise & Haedoch'i & Kim Dongin \\
47 & 화환 & Floral Tribute & Hwahwan & Lee Sang \\
48 & 황소와 도깨비 & The Bull and the Goblin & Hwangso-wa Tokkaebi & Yun Baengnam \\
49 & 후백제비화 & Later Baekje Chronicles & Hubaekjebihwa & I Iksang \\
50 & 흙의 세례 & Baptism of Soil & Hŭk-ŭi Selye & Hyeon Jingeon \\

\bottomrule
\end{tabular}}
\caption{List of novels included in the KoCoNovel Dataset. The romanization of authors' names is according to The Revised Romanization of Korean.}
\end{table*}

\end{document}